\documentclass[10pt,twocolumn,letterpaper]{article}

\usepackage[cvprfinal]{cvpr}
\usepackage{placeins}

\definecolor{cvprblue}{rgb}{0.21,0.49,0.74}
\usepackage[pagebackref,breaklinks,colorlinks,allcolors=cvprblue]{hyperref}

\def\paperID{ReGenAI-7} 
\def\confName{CVPR}
\def\confYear{2025}

\title{Dynamic watermarks in images  generated by diffusion models}

\author{
Yunzhuo Chen$^{1}$, Jordan Vice$^{1}$, Naveed Akhtar$^{2,1}$, Nur Al Hasan Haldar$^{3,1}$, Ajmal Mian$^{1}$ \\
$^{1}$The University of Western Australia, Perth, Australia \\
$^{2}$The University of Melbourne, Melbourne, Australia \\
$^{3}$Curtin University, Perth, Australia \\
{\tt\small yunzhuo.chen@research.uwa.edu.au, naveed.akhtar1@unimelb.edu.au,} \\
{\tt\small nur.haldar@curtin.edu.au, jordan.vice@uwa.edu.au, ajmal.mian@uwa.edu.au}
}

\begin{document}
\maketitle

\begin{abstract}

High-fidelity text-to-image diffusion models have revolutionized visual content generation, but their widespread use raises significant copyright concerns. We propose a novel multi-stage watermarking framework for diffusion models, designed to establish copyright and trace generated images back to their source. Our technique involves embedding: (i) a fixed watermark that is localized in the diffusion model's learned noise distribution and, (ii) a human-imperceptible, dynamic watermark in generates images, leveraging a fine-tuned decoder. By leveraging the Structural Similarity Index Measure (SSIM) and cosine similarity, we adapt the watermark's shape and color to the generated content while maintaining robustness. We demonstrate that our method enables reliable source verification through watermark classification, even when the dynamic watermark is adjusted for content-specific variations. Source model verification is enabled through watermark classification. We generate a dataset of watermarked images and introduce a methodology to evaluate the statistical impact of watermarking on generated content. Additionally, we rigorously test our framework against various attack scenarios, demonstrating its robustness and minimal impact on image quality. 
\end{abstract}    
\vspace{-4mm}
\section{Introduction}
\label{sec:intro}
\vspace{-2mm}
Recent deep generative models have made photorealistic image generation more accessible \citep{binkowski2018demystifying, heusel2017gans, salimans2016improved, zhou2019hype}, as exemplified by DALL·E 2~\citep{ramesh2022hierarchical}, Stable Diffusion~\citep{rombach2022high}, and FLUX models~\citep{FLUXAI2024}. These models have also spurred the development of numerous image editing tools and text-to-video models, including ControlNet~\citep{zhang2023adding}, Instruct-Pix2Pix~\citep{brooks2022instructpix2pix}, and SORA~\citep{Brooks2024}. While these models represent valuable digital assets, their widespread use raises the potential for misuse in spreading false narratives, generating harmful representations, or infringing intellectual property (IP) rights.

To address these challenges, blind watermarking has emerged as a key strategy for IP protection and misuse prevention. However, current watermarking methods face limitations, as fixed watermarks embedded during training are often predictable and vulnerable to removal or tampering through reverse engineering or image-processing techniques~\citep{moffat2019huffman, binkowski2018demystifying, heusel2017gans, salimans2016improved, zhou2019hype}. We propose a novel dual watermarking strategy by embedding: (\textit{i}) a unique, model-specific QR-code watermark directly into the diffusion model, and (\textit{ii}) dynamic watermarks into images generated by the model. The fixed QR-code watermark uniquely encodes model-specific metadata, including the IP address of the training machine and timestamp information.

 Our dynamic watermarking process enhances robustness by dynamically adjusting watermark transformations based on generated content within both feature and pixel spaces. In the feature space, high-level representations are extracted from original and watermarked images using a pre-trained feature extractor. We then calculate cosine similarity~\citep{rahutomo2012semantic} between these features to quantify their semantic consistency. In the pixel space, image quality is evaluated using the Structural Similarity Index (SSIM)~\citep{afchar2018mesonet, chen2022deepfake, cozzolino2017recasting, do2005contourlet, chen2024statistical}. We propose a validation method that assesses the impact of blind watermarking on image quality through an analysis of 11 distinct image statistics.

\vspace{-2mm}
\section{Related Work}
\vspace{-2mm}

\begin{figure*}
\centering
  \includegraphics[width=0.8\textwidth]{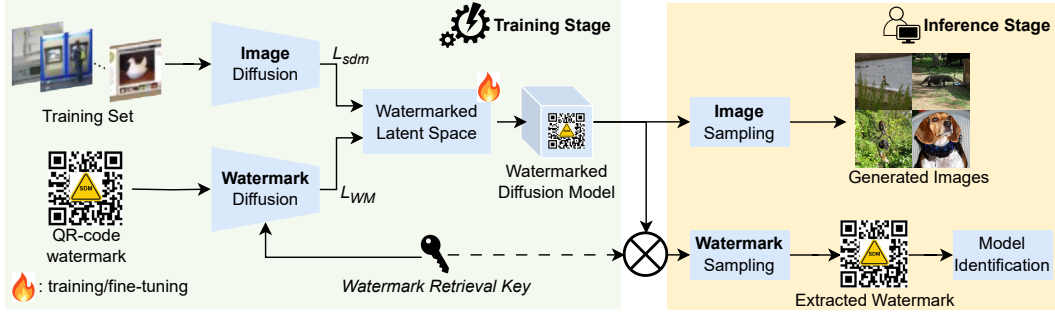}
\caption{We embed a QR-code watermark into a target diffusion model, leveraging a watermark retrieval key to isolate watermark data and image data distributions. \textbf{(Left)}  The training stage contains both watermark and image diffusion stages to construct a common, watermarked latent space. \textbf{(Right)}  Sampling from the watermarked model will generate images as usual. When provided with the watermark retrieval key as an input in the sampling process, the watermark will be extracted, allowing for model identification}

\label{2}       
\end{figure*}

Fundamentally, diffusion models are generative frameworks inspired by diffusion processes in non-equilibrium thermodynamics. Typically, these models implement forward and reverse diffusion processes via finite Markov chains~\citep{geyer1992practical}. Recently, diffusion models have been adapted for conditional image generation tasks, such as image inpainting, text-guided generation, and editing. Their iterative denoising steps enable zero-shot image editing by guiding the generative process~\citep{heusel2017gans, salimans2016improved, zhou2019hype}.

In generative models, watermark embedding within training datasets has been explored to protect the model's data~\citep{chai2020what}. However, this method can be inefficient, as adding new watermark information typically requires costly retraining. Alternative approaches integrate watermark embedding directly into the generative process~\citep{gragnaniello2021are, wang2020cnn}, closely aligning with model watermarking techniques. Yet, such methods have two primary limitations: (\textit{i}) they predominantly target GAN-based models, despite latent diffusion models (LDMs) increasingly replacing GANs in practical applications; and (\textit{ii}) watermarks are embedded during initial model training~\citep{fei2022supervised, lin2022cycleganwm}, which is resource-intensive and difficult to maintain. Recent studies indicate that efficient watermark embedding can be achieved through optimized fine-tuning of the generative model's latent decoder combined with an effective watermark extractor~\citep{fernandez2023stable}.
\vspace{-2mm}
\section{Proposed Method}
\vspace{-2mm}
We propose two watermark embedding branches: (\textit{i}) a \textit{model} watermarking branch that embeds a fixed QR-code watermark into the learned noise distribution of the diffusion model, ensuring robust traceability and ownership verification; and (\textit{ii}) an \textit{image} watermarking branch that embeds dynamic watermarks into generated images, balancing imperceptibility with resilience to attacks.
\vspace{-1mm}
\subsection{Model Watermarking}
\vspace{-1mm}

We employ the Stable Diffusion Model (SDM)~\citep{rombach2022high} for our experiments. To enable watermark embedding, we modify the latent diffusion process by introducing a modified Gaussian kernel. The data distribution is defined as \( q(z) \), where \( z \) represents data in the latent space. The watermark diffusion process serves as an extension of the traditional SDM diffusion process, introducing a watermark retrieval \textit{key}, which we denote as `$\kappa$' (see Fig. \ref{2}) to alter the diffusion pathway for the variable \( z_t \) such that:
\vspace{-2mm}
\begin{equation}
\hat{z}_t = \gamma_\kappa z_t + (1 - \gamma_\kappa)\kappa,
\end{equation}

where \(\gamma_\kappa\) is a blending factor that modulates the influence of the watermark on the generated output. Watermark embedding can be accomplished by fine-tuning the host model \( \epsilon_{\theta}^o \). In each iteration, we sample a data instance \( z_0 \) from the training dataset \( D_{\text{train}} \) and a watermark example \( z_0^{\text{w}} \) from the watermark dataset \( D_{\text{wm}} \). Noise samples \( \epsilon \) and \( \epsilon_{\text{w}} \) are then drawn from \( N(0, I) \) separately for task and watermark data, along with a timestep \( t \) sampled from \( \text{Uniform}(\{1, ..., T\}) \).

For the watermark sample \( z_0^{\text{w}} \), we first compute its latent representation \( z_t^{\text{w}} \) at timestep \( t \) within the latent diffusion process, and then construct the state \( \tilde{z}_t^{\text{w}} \) in the Watermark Diffusion Process based on \( z_t^{\text{w}} \) as follows:
\vspace{-2mm}
\begin{equation}
\tilde{z}_t^{\text{w}} = \gamma_\kappa\left( \sqrt{\bar{\alpha}_t} z_0^{\text{w}} + \sqrt{1 - \bar{\alpha}_t} \epsilon_{\text{w}} \right) + (1 - \gamma_\kappa) \kappa.
\end{equation}

 The joint learning optimization objective for the latent and watermark diffusion processes, which also serves as the loss function for watermark embedding:
\vspace{-2mm}
\begin{equation}
\begin{split}
L_{WDP} = {E}_{t \sim [1, T], z_0, z_{w0}, \epsilon_t} \Big[ & \, \gamma_{\epsilon}\| \epsilon - \epsilon_\theta(z_t, t) \|^2 \\
& + \| \epsilon_w - \epsilon_\theta(\hat{z}_{wt}, t) \|^2 \Big],
\end{split}
\end{equation}

Here, \( \gamma_{\epsilon} \) is a weighting factor that balances the standard diffusion process \( \epsilon \) and the watermark diffusion process \( \epsilon_w \).

\begin{figure*}
\centering
  \includegraphics[width=1\textwidth]{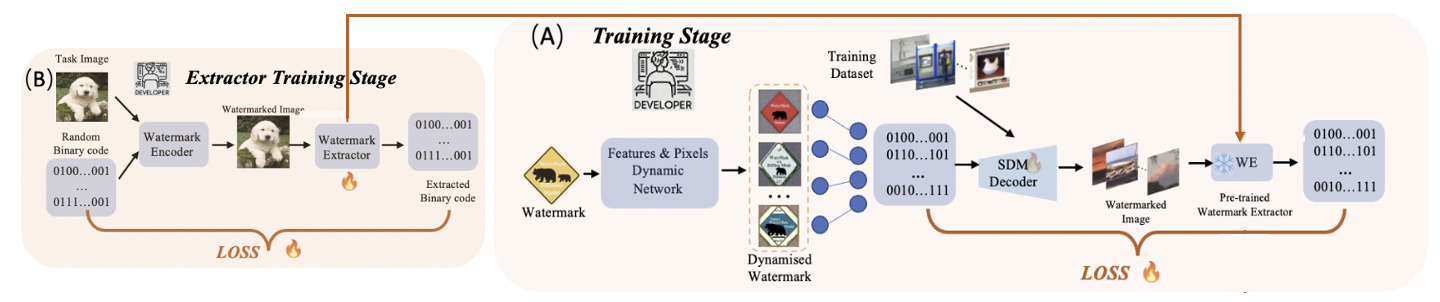}
\caption{Overview of the dynamic watermarking framework, structured in three stages: (A) Training Stage: Fine-tunes the SDM decoder with a pre-trained watermark extractor to embed dynamic watermarks. (B) Extractor Training Stage: Trains the watermark extractor to encode and retrieve binary watermark information, maintaining reliability under transformations}
\label{3}       
\end{figure*}

Watermark extraction can be achieved using the standard reverse diffusion process given the common latent space that was constructed through the combined diffusion processes discussed previously. The resulting output should be a reconstruction of the QR-code watermark that was used in the joint-training process.Given a model \( \epsilon_{\theta} \), along with the known trigger \( \kappa \) and trigger factor \( \gamma_\kappa \), we first sample \( z_T^{\text{w}} \) from \( N(0, 1) \). Next, we calculate its corresponding state \( \tilde{z}_t^{\text{w}} \) in the watermark diffusion process as follows: 
\vspace{-2mm}
\begin{equation}
\tilde{z}_t^{\text{w}} = \gamma_\kappa z_t^{\text{w}} + (1 - \gamma_\kappa)\kappa.
\end{equation}

This state \( \tilde{z}_t^{\text{w}} \) is then used as input to the model \( \epsilon_{\theta} \) to obtain the shared reverse noise to compute \( z_{t-1}^{\text{w}} \). The extraction process follows the formula:
\vspace{-2mm}
\begin{equation}
\begin{split}
{z}_{t-1}^{\text{w}} &= \frac{1}{\sqrt{\alpha_t}} \left[ {z}_t^{\text{w}} - \frac{(1 - \alpha_t)}{\sqrt{1 - \bar{\alpha}_t}} \epsilon_{\theta}\Bigl(\gamma_\kappa \hat{z}_t^{\text{w}} + (1 - \gamma_\kappa) \kappa, t\Bigr) \right] \\
&\quad + \sigma_t z,
\end{split}
\end{equation}

\vspace{-4mm}
where \( \sigma_t \) is the noise level at timestep \( t \) and \( z \) is a random sample from \( N(0, I) \). The selection of the trigger \( key \) and the factor \( \gamma_\kappa \) should ensure sufficient divergence between the state distribution in the watermark diffusion process and that in the standard diffusion process.

\vspace{-1mm}
\subsection{Image Watermarking}
Watermarks vary in shape and color while preserving key features of the reference watermark. In the feature space, we minimize the cosine similarity between original and transformed watermark features, enhancing orthogonality and robustness against extraction or tampering. Simultaneously, in the pixel space, we maximize the Structural Similarity Index Measure (SSIM) between original and transformed watermark images, ensuring high visual fidelity. These strategies enhance watermark robustness, imperceptibility, and classifiability without compromising generative capabilities.

To jointly satisfy constraints in both feature and pixel spaces, we formulate an optimization problem with the following objective:
\vspace{0mm}
\begin{equation}
\min_{} \left( \lambda_{cosine} \cdot \cos \theta - \lambda_{SSIM} \cdot \text{SSIM}(I_o, I_t) \right),
\end{equation}

where \(\lambda_{cosine}\) and \(\lambda_{SSIM}\) are weighting parameters that control the relative importance of cosine similarity and SSIM in the objective function. Original and transformed watermark images denoted as \(I_o\) and \(I_t\).

 In Fig. ~\ref{3} (A), the watermark image is first converted into a binary sequence. We employ Huffman coding-based character encoding to reduce the size of binary sequence~\cite{gajjala2020huffman}.  Watermark embedding consists of two steps. First, we train a watermark extractor. Then, we fine-tune the SDM decoder to embed specific watermark information in all generated images. 

As shown in Fig.~\ref{3}(B), the watermark extractor \( W_E \) training builds on HiDDeN~\citep{hsu1999hidden}. We embed binary watermark information by jointly optimizing the parameters of the watermark encoder and \( W_E \). Only \( W_E \) is retained as the watermark extractor. The watermark encoder takes a training image \( x_o \) and a binary watermark message \( m \) as inputs, producing a residual image \( x_\delta \). The watermarked image is obtained by scaling \( x_\delta \) with a factor \( \alpha \). The extractor \( W_E \) then recovers the binary watermark \( m' \) as:
\vspace{-2mm}
\begin{equation}
m' = W_E(T(x_o + \alpha x_\delta))
\end{equation}
\vspace{-2mm}

In Fig.~\ref{3} Part (A), the SDM uses the latent vector decoded by decoder to produce a generated image. To support multiple watermarks, the decoder \( D_m \) is extended to accept both the latent vector \( z \) and the condition vector \( e_i \), which specifies which watermark to embed. A training image is combined with the embedding vector \( e_i \) (derived from the condition \( i \)) to control the watermark embedding as follows:
\vspace{-2mm}
\begin{equation}
x'_w = D_m(E(x) \in {R}^{h \times w \times c}, e_i)
\end{equation}

The pre-trained extractor network \( W_E \) recovers the watermark \( m'_i \) from the generated image \( x'_w \). The binary watermark information loss ensures that the extracted watermark \( m'_i \) matches the target \( m_i \) specified by the condition \( i \):
\vspace{-4mm}
\begin{align}
L_m = -\sum_{j=1}^{k} \Big[ & {m_i}^{(j)} \cdot \log \sigma\left({m'_i}^{(j)}\right) \nonumber \quad + \\
& \left(1 - {m_i}^{(j)}\right) \cdot \log\left(1 - \sigma\left({m'_i}^{(j)}\right)\right) \Big]
\end{align}
\vspace{-5mm}

\vspace{-2mm}
\section{Watermark Extraction Results}
\vspace{-2mm}

\begin{table*}
     \scriptsize
    \centering
    
   \setlength{\tabcolsep}{0.3mm} 
\renewcommand\arraystretch{1} 
    \caption{Image statistics comparison results. "Difference" refers to the percentage difference in data.}
    \label{Statistics}.
      \resizebox{\textwidth}{!}{
    \begin{tabular}{c|c|c|c|c|c|c|c|c|c|c|c}
   
     & GLCM  & GLCM  & Canny  & Variance Blur & Mean  & Edge  & Entropy & Sharpness & Saturation & Texture   &  Image \\ 
     & Contrast & Energy & Edge & Measure & spectrum & Histogram & Strength  &  Score& & & Realism  \\
     
    \cline{1-12}
    Watermarked & 354.80 & 0.19 & 46.41 & 1124.47 & 85.73 & 6.00 & 6.31 & 8836.29  & 63.04 & 5.92  & 1.55\\
    Clean       & 371.23 & 0.19 & 48.34 & 1209.43 & 88.18 & 7.00 & 6.23 & 8903.23  & 67.00 & 5.65  & 1.42 \\
    Difference  & 4.63\% & 0.00\% & 4.16\% & 7.56\% & 2.86\% & 16.67\% & 1.27\% & 0.76\%  & 6.28\% & 4.56\%   & 8.39\%\\
  
    \end{tabular}}
\end{table*}

\begin{table*}
    \centering
    \scriptsize
    \renewcommand\arraystretch{1}
    \setlength{\tabcolsep}{1mm}{
    \caption{Classification results of Watermarked Images under various attacks. The classification network distinguishes whether an image contains a watermark and then classifies the watermark.}
    \label{tab:attacks}
    \resizebox{\linewidth}{!}{%
    \begin{tabular}{l|c|c|c|c|c|c|c}
   
     \multicolumn{2}{c}{} & \multicolumn{6}{c}{Attack Type} \\
    \cline{2-8}
  
    & No Attack & Rotation & Blurring  & Texture Reduction & Image Compression & Crop & Flip \\
    \cline{1-8}
     
    Watermark Presence & 100.00\% & 99.30\% & 100.00\% & 95.70\% & 98.50\% & 99.10\% & 100.00\% \\
    \cline{1-8}
    Watermark Classification & 97.00\% & 95.98\% & 93.97\% & 93.94\% & 94.72\% & 94.05\% & 96.10\% \\
    
    \end{tabular}}}
    \vspace{-0mm}
\end{table*}

\begin{figure}[h]
\centering
\includegraphics[width=0.45\textwidth]{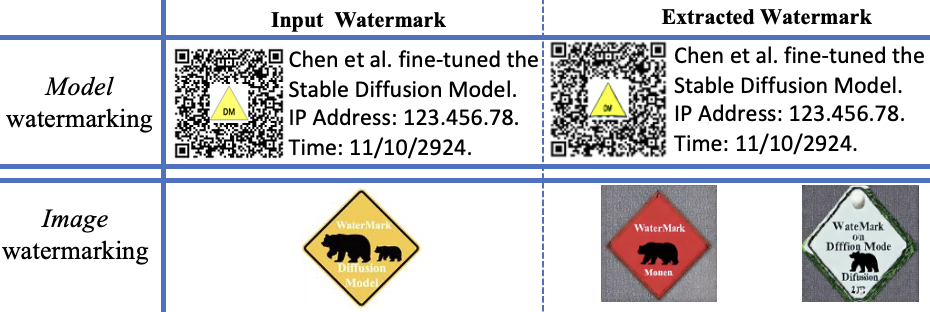}
\caption{Raw 1 shows the extracted QR code watermark from the model watermark embedding branch. Raw 2 shows extracted watermarks from the image watermark embedding branch}
\label{fig:model_watermark}
\end{figure}

In  row 1 of Fig.~\ref{fig:model_watermark}, our method effectively preserves the watermark’s visibility and structure after extraction. The central logo undergoes minor deformation due to the diffusion process's inherent noise and transformations, which does not impact the QR code’s scalability or embedded information. In row 2, our custom-designed "Diffusion Model" and "Watermark" serve as the core information. Each extracted watermark retains this text, while dynamic transformations modify its appearance. These transformations enhance robustness and resilience, yet the classifier can still accurately verify the watermark’s origin.

We use image statistics based on human perception as evaluation metrics. Based on the sensitivity coefficient proposed by ~\citep{chen2024statistical}, we selected 11 image statistics measures with sensitivity values above 0.3, indicating a high sensitivity to changes in texture, edges, and frequency. Sensitivity ranges from 0 to 1, with higher values indicating greater sensitivity. As shown in Table~\ref{Statistics}, the average change between watermarked and clean images across these metrics is  minimal, approximately 5\%.

As shown in Table~\ref{tab:attacks},  texture reduction caused the largest drop (4.3\%), likely due to its impact on watermark details, weakening detectability. Image compression, particularly lossy compression, also reduced accuracy to 98.50\%. However, flip and crop attacks had minimal impact, maintaining accuracy at 100.00\% and 99.10\%, respectively. For the "Watermark Classification" task, attacks slightly reduced classification accuracy. Blurring and texture reduction had the greatest impact, lowering accuracy to 93.97\% and 93.94\%, respectively, suggesting these attacks degrade visual features, making classification more challenging.

\begin{table}[h]
    \centering
    \caption{Comparison of Image Generation Quality using IS and FID}
    \label{tab:comparison}
    \begin{tabular}{l|c|c}
        \hline
        Method & IS $\uparrow$ & FID $\downarrow$ \\
        \hline
        Tree-Ring \cite{wen2023tree} & 3.80 & 3298.22 \\
        Chen et al. \cite{chen2025image} & 4.43& 2789.98 \\
        Xin et al. \cite{li2017simplified} & 4.14 & 3477.38 \\
        Clean images & 5.01 & 2157.45 \\
        \textbf{Ours} & \textbf{4.61} & \textbf{2687.34} \\
        \hline
    \end{tabular}
\end{table}

To evaluate the effectiveness of our watermark embedding method in image generation, we compare our approach with three existing watermark embedding techniques using two widely used metrics: Inception Score (IS)~\cite{salimans2016improved} and Fréchet Inception Distance (FID)~\cite{heusel2017gans}. The IS measures the diversity and realism of the generated images, where higher scores indicate better quality. The FID assesses the distribution similarity between generated and real images, with lower scores indicating better quality. The comparison results are presented in Table~\ref{tab:comparison}. Our method outperforms the other approaches, achieving a higher IS and a lower FID, demonstrating its superior performance in maintaining image quality while embedding watermarks.

\vspace{-3mm}
\section{Conclusion}
\vspace{-2mm}
In this paper, we propose a dual approach for embedding fixed QR-codes within the diffusion process and dynamic watermarks in generated images. This integration enhances intellectual property protection and traceability in generated content. The dynamic watermark undergoes feature and pixel space transformations, increasing resistance to attacks while preserving image quality.  It remains intact even under rotation, blurring, and compression. Statistical validation shows dual watermarking method has minimal impact on image quality, with only a 5\% variation in key metrics such as edge and texture attributes, confirming the method's invisibility and robustness. Our work addresses critical challenges in the ethical use of AI-generated content, providing a scalable and effective mechanism for ownership verification and misuse prevention. Future research could explore extending this framework to other generative models and applications, further advancing the field of digital security.

\section{Acknowledgement}
This research was partially supported by National Intelligence and Security Discovery Research Grants (project NS220100007), funded by the Department of Defence, Australia

{
    \small
    \bibliographystyle{ieeenat_fullname}
    \bibliography{main}
}


\clearpage

\setcounter{page}{1}
\section{Supply Material }

\subsection{ More Watermark Extraction Results}
\label{1}

\begin{figure}[h]
\centering
\includegraphics[width=0.5\textwidth]{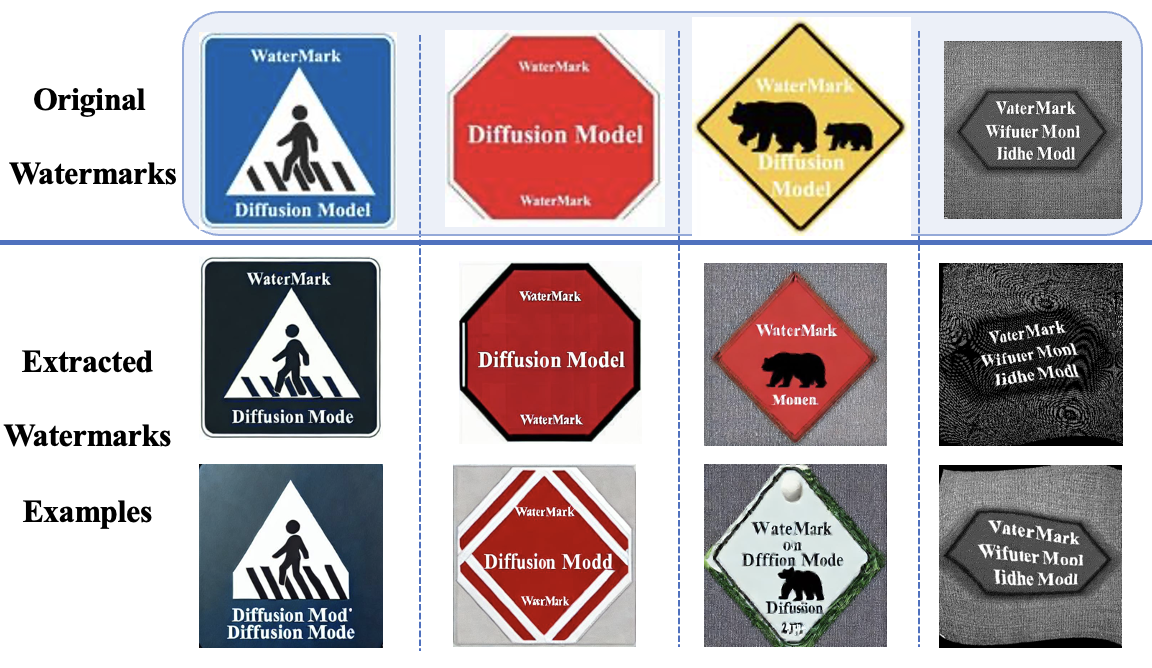}
\caption{Extracted watermarks from the image watermark embedding branch. Each watermark has been dynamically transformed in shape and color.}
\label{fig:image_watermark}
\end{figure}

The watermark’s visibility and structure after extraction. In Fig.~\ref{fig:model_watermark}, our method effectively preserves the watermark’s visibility and structure after extraction. The central logo undergoes minor deformation due to the diffusion process's inherent noise and transformations, which does not impact the QR code’s scalability or embedded information. In Fig.~\ref{fig:image_watermark},, our custom-designed "Diffusion Model" and "Watermark" serve as the core information. Each extracted watermark retains this text, while dynamic transformations modify its appearance. These transformations enhance robustness and resilience, yet the classifier can still accurately verify the watermark’s origin.

\end{document}